%% file: conference_101719.tex
\documentclass[conference]{IEEEtran}
\IEEEoverridecommandlockouts

\usepackage{cite}
\usepackage{amsmath,amssymb,amsfonts}
\usepackage{algorithmic}
\usepackage{multirow}
\usepackage{graphicx}
\usepackage{textcomp}
\usepackage{xcolor}
\usepackage{orcidlink}
\usepackage{hyperref}
\usepackage{url}
\usepackage{csquotes}
\usepackage{booktabs}
\usepackage{makecell}      
\usepackage{siunitx}       
\usepackage{amsmath}
\usepackage{eurosym}       
\usepackage{soul}
\usepackage{rotating} 
\def\BibTeX{{\rm B\kern-.05em{\sc i\kern-.025em b}\kern-.08em
    T\kern-.1667em\lower.7ex\hbox{E}\kern-.125emX}}

\begin{document}

\title{Energy Efficiency of Locally Deployed LLMs:\\ A Preliminary Quantitative GPU Power Benchmark on Consumer Hardware
}

\author{
\IEEEauthorblockN{Philipp M. Zähl~\orcidlink{0000-0003-3302-4415}, Elja Dalipaj, Anika Hennig~\orcidlink{0009-0001-9418-8512}} and Timon Bayer~\orcidlink{0009-0000-4589-0287}
\IEEEauthorblockA{\textit{Institute for Digitalization Aachen} \\
\textit{FH Aachen University of Applied Sciences}\\
Aachen, Germany \\
\{zaehl, dalipaj, hennig, t.bayer\}@fh-aachen.de}
}

\maketitle

\begin{abstract}
The local deployment of large language models (LLMs) is gaining traction due to privacy concerns and the desire for on-premise inference. However, the energy costs on consumer hardware remain poorly characterized, as most benchmarks focus solely on accuracy. This paper presents a reproducible, hardware-level energy benchmark of 18 open-source LLMs (0.5B to 7B parameters) executed on a single consumer GPU (RTX~4060~Ti 16GB). Using the Ollama inference engine, GPU power draw was sampled at \SI{2}{\hertz} via \textit{nvidia-smi} across a fixed prompt set. We evaluate mean/peak power, total energy per prompt (J/prompt), energy per output token (J/token), and throughput (tok/s). Our findings suggest that factors beyond raw parameter count, including model architecture and quantization strategy, drive energy efficiency. Specifically, \texttt{qwen2.5:0.5b} and \texttt{tinyllama:1.1b} achieve the lowest energy cost (\SI{0.2747}{\joule\per token} and \SI{0.3234}{\joule\per token}) and the highest throughput ($>$\SI{325}{tok\per s}). In contrast, the 7B-Mistral model consumes up to $8.6\times$ more energy per token than the most efficient model. Notably, \texttt{qwen3.5:0.8b(on)} exhibits anomalously high per-prompt energy due to extended internal reasoning, highlighting the need to distinguish between token generation modes in efficiency metrics.
\end{abstract}


\input{text}

\section*{Supplementary Material} \label{sec_suppleMat}
Please visit the following URL for our code, the prompt set, plots, and raw data: \url{https://github.com/fhac-ida/GreenCode-LLM-Benchmark}

\section*{Acknowledgement}
This work was created in the context of the ITEA4 and Eureka Cluster on the software innovation project "23016 GreenCode", supported by funding from the Federal Ministry of Research, Technology and Space (BMFTR), Germany (grant number 16IS24070G).

\bibliographystyle{IEEEtran}
\bibliography{literatur_new}

\end{document}

%% file: text.tex
\section{Introduction}
  
The democratization of large language model inference through tools such as Ollama~\cite{ollama2023}, llama.cpp~\cite{llamacpp2023}, and LM~Studio~\cite{lmstudio2024} has made it feasible for individual users and small organizations to run state-of-the-art language models locally, without transmitting data to third-party cloud providers \cite{aminabadi2022deepspeed}. Locally deployed models offer advantages in latency, data privacy, and long-term costs, but they also impose a continuous energy burden on the host machine that, at scale, is non-trivial \cite{argerich2024measuring, patterson2021carbon, poddar2025sustainable}.
 
The environmental impact of large-scale AI training has been well documented \cite{strubell2019,patterson2021carbon}, but inference energy, especially on consumer hardware, is significantly less studied. For an end user running a 7B-parameter model as a daily assistant, the cumulative energy cost over months of use can rival that of other household appliances \cite{argerich2024measuring, patterson2021carbon, eurostat2024}. As model proliferation accelerates and edge-deployment tooling matures, understanding the energy-performance tradeoffs of available models becomes a practical engineering concern.
 
This paper addresses the following research question: \textit{Among openly available language models deployable on a single consumer GPU, which models provide the most favorable energy efficiency, and how do efficiency metrics relate to model size and architecture}? Therefore, our contributions are fourfold: First, we provide a reproducible, open-source benchmark harness that instruments GPU power draw via \texttt{nvidia-smi} for containerized LLM inference workloads. Second, we contribute a diverse empirical baseline covering multiple energy and performance metrics across several open-source model families. Third, we provide a systematic analysis of the relationship between model size and energy efficiency, demonstrating how architectural scaling choices affect throughput and operational costs even when accounting for response quality. Finally, we offer new insights into how advanced prompting and reasoning mechanisms, such as extended chain-of-thought (CoT), fundamentally alter the established energy profiles of smaller model classes.
 
The remainder of this paper is organized as follows: First, Section~\ref{sec:related} reviews related work. Then, Section~\ref{sec:methodology} describes the experimental setup and measurement protocol. Section~\ref{sec:results} presents results, followed by a discussion on the implications in Section~\ref{sec:discussion}. Finally, we conclude the paper in Section~\ref{sec:conclusion}.

\section{Related Work}
\label{sec:related}

\subsection{Energy Cost of AI Workloads}
 
Strubell~et~al.~\cite{strubell2019} first brought attention to the substantial carbon footprint of training large natural language processing (NLP) models, reporting that training a single Transformer-based NLP model could emit as much CO\textsubscript{2} as five transatlantic flights. Patterson~et~al.~\cite{patterson2021carbon} extended this analysis to include inference and found that, in production deployments, inference energy often dominates lifetime energy costs. Schwartz~et~al.~\cite{schwartz2020greenai} coined the term ``Green~AI'' and called for efficiency metrics to be reported alongside accuracy in research publications. \text{Argerich and Patiño-Martínez}~\cite{argerich2024measuring} reinforced this shift by arguing that since models are trained once but queried millions of times in production, the cumulative inference footprint quickly surpasses training costs, as demonstrated by the massive daily energy demands of real-world conversational agents.
 
\subsection{LLM Inference Benchmarking}
 
Several recent efforts have benchmarked LLM inference performance. Aminabadi~et~al.~\cite{aminabadi2022deepspeed} characterized the throughput and latency of large models under different parallelism strategies on server-class hardware. Frantar~et~al.~\cite{frantar2022gptq} and Dettmers~et~al.~\cite{dettmers2023qlora} demonstrated that quantization dramatically reduces VRAM requirements and can improve inference speed, though its energy implications have not been systematically quantified for consumer deployments. \text{Poddar et al.}~\cite{poddar2025sustainable} addressed this gap through an extensive benchmarking of both encoder-decoder and decoder-only architectures across diverse NLP tasks, establishing that inference energy is more strongly correlated with output token length and response time than input characteristics. To further capture query-level variability, \text{Maliakel et al.}~\cite{maliakel2025dvfs} demonstrated that surface-level heuristics like input length fail to predict computational difficulty, which is instead better explained by lightweight semantic features such as entity density.
 
\subsection{Measurement Methodology}
 
Bannour~et~al.~\cite{bannour2021evaluating} surveyed tools for measuring the carbon footprint of NLP pipelines, including \texttt{codecarbon} \cite{codecarbon2021} and Intel RAPL-based utilities. GPU-level measurement via \texttt{nvidia-smi} has been widely adopted in practice due to its availability on all NVIDIA-based systems, though prior work has noted a polling granularity limitation \cite{pynvml2020}. To overcome the challenge of profiling isolated hardware components, \text{Argerich and Patiño-Martínez}~\cite{argerich2024measuring} introduced a pure software framework capable of granularly isolating CPU, memory, GPU, and storage energy draw on Linux systems. At the micro-architectural phase level, \text{Maliakel et al.}~\cite{maliakel2025dvfs} leveraged NVML telemetry to separate the compute-bound prefill phase from the memory-bound autoregressive decode phase. They discovered that the decode phase dominates up to 91\% of inference time but is highly insensitive to GPU core frequency, meaning that phase-aware dynamic voltage and frequency scaling (DVFS) can yield up to 42\% energy savings with negligible latency penalties.

\subsection{Analytical Energy Estimation}

While the studies above rely on direct hardware telemetry, a complementary work estimates inference energy analytically when runtime instrumentation is unavailable or impractical. Vartziotis et al. \cite{vartziotis2026tokens} propose a semi-analytical GPU-level estimator that decomposes request energy into tensor-core compute and HBM memory-movement terms, calibrated with hardware-specific coefficients for H100-class accelerators. Their approach separates prefill from autoregressive decode and models attention-related KV-cache traffic as scaling super-linearly with output length - a scaling behavior consistent with the throughput degradation we observe empirically for longer, reasoning-augmented generations (Section~\ref{sec:discussion}). However, their estimator targets datacenter-class hardware and large parameter-count regimes and is explicitly not intended to replace measurement; it relies on published energy measurements for calibration rather than direct instrumentation. Our work is complementary: we provide ground-truth, hardware-measured energy data for the small-model, consumer-GPU field that such analytical estimators do not currently cover and which could in principle serve as an independent calibration point for extending their methodology to non-datacenter accelerators.

\section{Methodology}
\label{sec:methodology}
 
\subsection{Testbed Hardware and Software}
 
All experiments were conducted on a single workstation running Ubuntu~24 under Windows Subsystem for Linux 2 (WSL2). Table~\ref{tab:hardware} summarizes the hardware configuration. The host GPU, an NVIDIA GeForce RTX~4060~Ti with \SI{16}{\gibi\byte} GDDR6X VRAM and CUDA Compute Capability~8.9, was used exclusively for inference; no concurrent GPU workloads were active during measurement.
 
\begin{table}[htb]
 \renewcommand{\arraystretch}{1.15}
 \caption{Experimental Hardware Configuration}
 \label{tab:hardware}
 \centering
 \begin{tabular}{ll}
  \toprule
  \textbf{Component} & \textbf{Specification} \\
  \midrule
  GPU        & NVIDIA GeForce RTX 4060 Ti \\
  VRAM        & \SI{16}{\gibi\byte} GDDR6X \\
  GPU Compute Cap.  & 8.9 (Ada Lovelace) \\
  GPU Driver     & 591.86 \\
  CPU        & Intel i7-12700K (Cores: 10/20) \\
  Subsystem RAM   & \SI{15.49}{\gibi\byte} (\SI{32}{\gibi\byte} available) \\
  Subsystem OS    & Ubuntu 24.04.4 LTS (WSL2, Windows 11) \\
  Inference Engine  & Ollama (Docker v4.69.0, GPU passthrough) \\
  \bottomrule
 \end{tabular}
\end{table}
 
Models were served by Ollama running inside an official Docker container (\texttt{ollama/ollama}) with full GPU passthrough (\texttt{--gpus all}). The Ollama API was accessed via HTTP at \texttt{localhost:11434} using the \texttt{/api/generate} endpoint in non-streaming mode.
 
\subsection{Models Under Test}
 
Eighteen openly available language models were selected to span a range of parameter sizes (1B-7B), model families, and quantization levels representative of the current Ollama model library. The main constraint in selecting a model was memory requirements: preference was given to models or model variants that require \SI{8}{\gibi\byte} or less of memory. Table~\ref{tab:models} lists the evaluated models. All models were obtained via \texttt{ollama pull} and use the default quantization shipped by the respective Ollama registry entry (typically 4-bit or 8-bit GGUF/AWQ quantization). Note on Reasoning Modes: Some models (e.g., \texttt{deepseek-r1:*}, \texttt{qwen3.5:0.*b (on/off)}) include configurable reasoning modes, where \texttt{(on)} enables extended chain-of-thought (CoT) or internal reasoning steps, and \texttt{(off)} disables them. These modes significantly alter energy consumption and latency, as reasoning-enabled models trade efficiency for improved response quality. All variants were benchmarked under their default Ollama quantization settings.

\begin{table*}[htb]
 \caption{Architectural and Operational Properties of Evaluated Models}
 \label{tab:models}
 \centering
 \small 
 \begin{tabular}{llcll}
  \toprule
  \textbf{Model Tag} & \textbf{Family} & \textbf{Params} & \textbf{Primary Type} & \textbf{Context} \\
  \midrule
  \texttt{deepseek-r1:1.5b} (off) & DeepSeek R1 & $\sim$1.5B & Reasoning & 128k \\
  \texttt{deepseek-r1:1.5b} (on) & DeepSeek R1 & $\sim$1.5B & Reasoning & 128k \\
  \texttt{deepseek-r1:7b} (off) & DeepSeek R1 & $\sim$7B & Reasoning & 128k \\
  \texttt{deepseek-r1:7b} (on) & DeepSeek R1 & $\sim$7B & Reasoning & 128k \\
  \texttt{gemma2:2b} & Google Gemma 2 & $\sim$2B & Multimodal / General & 128k \\
  \texttt{gemma3:1b} & Google Gemma 3 & $\sim$1B & Multimodal / General & 128k \\
  \texttt{gemma3:4b} & Google Gemma 3 & $\sim$4B & Multimodal / General & 128k \\
  \texttt{gemma4:e2b} (off) & Google Gemma 4 & $\sim$2B & Reasoning / Edge & 128k \\
  \texttt{gemma4:e2b} (on) & Google Gemma 4 & $\sim$2B & Reasoning / Edge & 128k \\
  \texttt{granite3.1-dense:2b} & IBM Granite 3.1 & $\sim$2B & Dense Model & 128k \\
  \texttt{llama3.2:1b} & Meta Llama 3.2 & $\sim$1B & General / Edge & 128k \\
  \texttt{llama3.2:3b} & Meta Llama 3.2 & $\sim$3B & General / Edge & 128k \\
  \texttt{mistral:7b} & Mistral & $\sim$7B & General & 32k \\
  \texttt{phi3:mini} & Microsoft Phi-3 & $\sim$3.8B & Reasoning & 128k \\
  \texttt{phi4-mini:3.8b} & Microsoft Phi-4 & $\sim$3.8B & Reasoning & 128k \\
  \texttt{qwen2.5:0.5b} & Alibaba Qwen 2.5 & $\sim$0.5B & General & 128k \\
  \texttt{qwen2.5:3b} & Alibaba Qwen 2.5 & $\sim$3B & General & 128k \\
  \texttt{qwen2.5:7b} & Alibaba Qwen 2.5 & $\sim$7B & General & 128k \\
  \texttt{qwen3.5:0.8b} (off) & Alibaba Qwen 3.5 & $\sim$0.8B & Reasoning / Vision & 262k \\
  \texttt{qwen3.5:0.8b} (on) & Alibaba Qwen 3.5 & $\sim$0.8B & Reasoning / Vision & 262k \\
  \texttt{qwen3.5:4b} (off) & Alibaba Qwen 3.5 & $\sim$4B & Reasoning / Vision & 262k \\
  \texttt{qwen3.5:4b} (on) & Alibaba Qwen 3.5 & $\sim$4B & Reasoning / Vision & 262k \\
  \texttt{tinyllama:1.1b} & TinyLlama & $\sim$1.1B & General & 2k \\
  \bottomrule
 \end{tabular}
\end{table*}
 
\subsection{Prompt Set and Execution Protocol}
 
A fixed set of 15 prompts was designed to span a representative range of real-world inference workloads, deliberately varying expected output length and computational demand. The prompts were organized into three tiers:
 
\begin{itemize}
 \item \textbf{Tier~1 - Short factual} (5 prompts): Single-sentence
    questions with brief, deterministic answers (e.g.,
    \emph{``What is the capital of France?''},
    \emph{``What is 17 multiplied by 23?''}).
    Expected output: 1-3 sentences.
 
 \item \textbf{Tier~2 - Medium reasoning} (5 prompts): Questions requiring
    explanation or multi-step synthesis
    (e.g., \emph{``Explain the difference between TCP and UDP in two
    sentences.''}, \emph{``What is gradient descent and why is it used in
    machine learning?''}).
    Expected output: 2-5 sentences.
 
 \item \textbf{Tier~3 - Long generation} (5 prompts): Open-ended generation
    tasks demanding structured, multi-paragraph responses
    (e.g., \emph{``Write a Python function that checks whether a given
    string is a palindrome. Include docstring and tests.''},
    \emph{``Explain how a transformer neural network works. Cover attention
    mechanisms, tokenisation, and training.''}).
    Expected output: 15-30 sentences.
\end{itemize}
 
This tiered design ensures that aggregate statistics are not dominated by either trivially short or unusually long completions and mirrors the heterogeneous workloads encountered in practice. The complete prompt file is included in the appendix. Each model processed the full 15-prompt set in three independent runs, yielding $15 \times 3 = 45$ prompt-level observations per model (\num{810} total). Three repetitions were selected as a pragmatic trade-off between experimental runtime and measurement robustness. Although some run-to-run variability was observed, it remained modest and did not alter the relative ranking or overall trends across models, indicating that three repetitions were sufficient for reliable comparative analysis.
 
For each model, the following protocol was applied:
\begin{enumerate}
 \item \textbf{Model pull}: The model weights were pre-downloaded and cached
    locally; the pull step only verified integrity.
 \item \textbf{Warmup}: A single throwaway prompt (\texttt{"hello"}) was
    issued to ensure the model weights were loaded into GPU VRAM before any
    measurement. An additional \SI{10}{\second} sleep followed.
 \item \textbf{Idle baseline}: GPU power was sampled for \SI{20}{\second}
    with the model resident in VRAM but idle. The mean idle power
    $P_\mathrm{idle}$ served as the baseline for delta-power calculations.
 \item \textbf{Load measurement}: The 15-prompt set was executed three times.
    For each prompt, GPU power sampling began immediately before the API
    call and ended upon response receipt.
 \item \textbf{Cooldown}: After all runs, the model was unloaded from VRAM
    via the Ollama \texttt{keep\_alive=0} API, and a \SI{15}{\second}
    cooldown was observed before loading the next model.
\end{enumerate}
This sequential, single-model design avoids interference between models and ensures that each model's idle baseline reflects the actual GPU idle state under Ollama's memory management.
 
\subsection{Power and Energy Measurement}

GPU power draw was queried using \texttt{nvidia-smi} at a nominal rate of \SI{2}{\hertz} (polling interval \SI{0.5}{\second}) running in a dedicated background thread. The \texttt{nvidia-smi} \texttt{power.draw} field reports the instantaneous board power in watts as provided by the NVML driver, which itself averages over a hardware counter window of approximately \SI{100}{\milli\second} \cite{pynvml2020}.
 
For each prompt-level inference call, the following quantities were derived:
 
\begin{align*}
 P_\mathrm{load,mean} &= \frac{1}{N}\sum_{i=1}^{N} P_i \\[4pt]
 E_\mathrm{GPU}    &= P_\mathrm{load,mean} \cdot \Delta t \\[4pt]
 \text{J/tok}     &= \frac{E_\mathrm{GPU}}{n_\mathrm{out}}
\end{align*}
 
where $P_i$ denotes the $i$-th power sample during the inference window, $\Delta t$ is the wall-clock duration of the inference call, $n_\mathrm{out}$ is the number of generated (output) tokens as reported by the Ollama API, and $N$ is the number of samples collected. Per-prompt results were aggregated as arithmetic means across the 45 observations per model.
 
Throughput (tok/s) was derived from the \texttt{eval\_count} and \texttt{eval\_duration} fields returned by the Ollama API, which measure only the autoregressive decoding phase, excluding the prefill stage.
 
\subsection{Reproducibility}
 
The complete benchmark harness, raw CSV timeseries, summary data, model specific charts and all figures are published in an open repository (see appendix).

\section{Results}
\label{sec:results}
 
Table~\ref{tab:summary} presents aggregate results across all evaluated models.
The following subsections discuss each metric dimension in turn.
 
\begin{table*}[!t]
 \renewcommand{\arraystretch}{1.15}
 \caption{Summary of Energy and Performance Metrics (mean over 45 prompt observations per model)}
 \label{tab:summary}
 \centering
 \begin{tabular}{l rrrr rrr}
  \toprule
  \textbf{Model} &
  \textbf{\makecell{Idle\\(W)}} &
  \textbf{\makecell{Load\\Mean (W)}} &
  \textbf{\makecell{Load\\Peak (W)}} &
  \textbf{\makecell{$\Delta P$\\(W)}} &
  \textbf{\makecell{Energy /\\Prompt (J)}} &
  \textbf{\makecell{J /\\Out-Token}} &
  \textbf{\makecell{Throughput\\(tok/s)}} \\
  \midrule
  \texttt{qwen2.5:0.5b}   & {18.25} & {63.47} & {100.70} & {45.22} & \textbf{98.58} & \textbf{0.275} & \textbf{329.0} \\
  \texttt{tinyllama:1.1b}   & 16.82 & 83.63 & 118.90 & 66.81 & 110.21 & 0.323 & 325.1 \\
  \texttt{gemma3:1b}   & 12.99 & 80.84 & 108.60 & 67.85 & 267.33 & 0.537 & 206.8 \\
  \texttt{deepseek-r1:1.5b (off)}   & 17.46 & 91.78 & 123.99 & 74.32 & 156.92 & 0.611 & 213.5 \\
  \texttt{qwen3.5:0.8b (off)}   & 19.95 & 86.24 & 101.56 & 66.29 & 319.31 & 0.621 & 169.6 \\
  \texttt{qwen3.5:0.8b (on)}   & 22.54 & 97.94 & 107.06 & 75.40 & 2244.17 & 0.640 & 171.4 \\
  \texttt{llama3.2:1b}   & 25.23 & 82.86 & 109.32 & 57.63 & 196.96 & 0.647 & 173.9 \\
  \texttt{deepseek-r1:1.5b (on)}   & 26.11 & 114.35 & 128.85 & 88.24 & 621.27 & 0.663 & 209.7 \\
  \texttt{granite3.1-dense:2b}   & 15.24 & 110.56 & 129.31 & 95.32 & 361.50 & 0.928 & 135.9 \\
  \texttt{gemma2:2b}   & 23.63 & 90.12 & 115.16 & 66.49 & 316.56 & 0.969 & 131.2 \\
  \texttt{gemma4:e2b (off)}   & 12.55 & 95.67 & 122.84 & 83.12 & 539.32 & 0.992 & 129.0 \\
  \texttt{gemma4:e2b (on)}   & 16.08 & 109.53 & 123.68 & 93.45 & 908.24 & 1.003 & 126.6 \\
  \texttt{qwen2.5:3b}   & 18.91 & 105.56 & 132.87 & 86.65 & 377.46 & 1.173 & 115.4 \\
  \texttt{phi3:mini}   & 14.79 & 113.94 & 124.40 & 99.15 & 534.81 & 1.196 & 103.4 \\
  \texttt{llama3.2:3b}   & 16.70 & 104.24 & 132.98 & 87.54 & 361.37 & 1.204 & 112.9 \\
  \texttt{phi4-mini:3.8b}   & 12.53 & 100.28 & 136.44 & 87.75 & 277.04 & 1.450 & 93.7 \\
  \texttt{gemma3:4b}   & 15.68 & 103.65 & 130.52 & 87.97 & 831.95 & 1.529 & 81.2 \\
  \texttt{qwen3.5:4b (off)}   & 12.34 & 113.78 & 128.70 & 101.44 & 878.74 & 1.866 & 69.3 \\
  \texttt{qwen3.5:4b (on)}   & 18.06 & 123.93 & 134.51 & 105.87 & 3589.98 & 1.940 & 67.5 \\
  \texttt{mistral:7b}   & 16.19 & 122.51 & 139.07 & 106.32 & 758.79 & 2.360 & 57.1 \\
  \texttt{deepseek-r1:7b (off)}   & 16.82 & 117.52 & 138.70 & 100.70 & 747.73 & 2.438 & 57.9 \\
  \texttt{deepseek-r1:7b (on)}   & 17.28 & 129.93 & 139.56 & 112.65 & 1838.71 & 2.475 & 57.6 \\
  \texttt{qwen2.5:7b}   & 19.92 & 118.65 & 138.68 & 98.73 & 820.84 & 2.478 & 57.5 \\
  \bottomrule
 \end{tabular}
 \vspace{1mm}

 \small Rows sorted by J/Out-Token (ascending). $\Delta P = P_\mathrm{load,mean} - P_\mathrm{idle}$.
 Best values per column are \textbf{bold}.
\end{table*}
 
\subsection{GPU Power Draw}
 
\begin{sidewaysfigure*} 
    \centering
    \includegraphics[width=0.95\textheight,trim={0cm 15cm 62cm 3.5cm},clip]{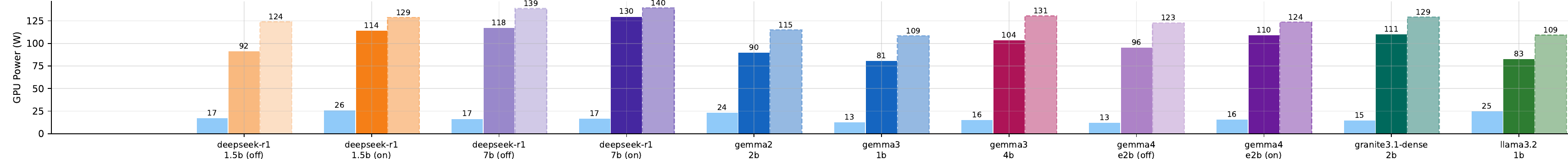}
    \includegraphics[width=0.95\textheight,trim={58cm 15cm 0cm 3.5cm},clip]{figures/comparison_bar_2.pdf}
    \caption{Idle/load/peak GPU power for all models.}
    \label{fig:gpu_power_general}
    \vspace{5mm}
    \includegraphics[width=0.95\textheight,trim={0cm 8cm 62cm 10.5cm},clip]{figures/comparison_bar_2.pdf}
    \includegraphics[width=0.95\textheight,trim={58cm 8cm 0cm 10.5cm},clip]{figures/comparison_bar_2.pdf}
    \caption{Average GPU energy per prompt for all models.}
    \label{fig:gpu_power_per_prompt}
    \vspace{5mm}
    \includegraphics[width=0.95\textheight,trim={0cm 0cm 62cm 18.5cm},clip]{figures/comparison_bar_2.pdf}
    \includegraphics[width=0.95\textheight,trim={58cm 0cm 0cm 18.5cm},clip]{figures/comparison_bar_2.pdf}
    \caption{GPU energy per output token for all models.}
    \label{fig:gpu_power_per_token}
\end{sidewaysfigure*}

Figure~\ref{fig:gpu_power_general} shows idle mean, load mean, and load peak GPU power for each model.
Idle power ranged from 12~W to 26~W, reflecting differences in VRAM occupancy and model keep-alive behavior.
During inference, mean load power spanned 63~W (\texttt{qwen2.5:0.5b}) to 130~W (\texttt{deepseek-r1:7b} (on)). The delta power $\Delta P$, defined as the inference-attributable power increment above the idle baseline, is reported to provide additional context for the measured energy consumption.
By separating the workload-related power draw from the system's static idle load, this metric helps interpret the power measurements shown in the figures (see Appendix) and provides a reference point for comparisons across different hardware platforms. In our measurements, $\Delta P$ ranged from 45~W for \texttt{qwen2.5:0.5b} to 113~W for \texttt{deepseek-r1:7b} (on), corresponding to an absolute difference of 68~W, or approximately 151\%.

Peak GPU power was more tightly clustered, ranging from 101~W (\texttt{qwen2.5:0.5b}) to 140~W (\texttt{deepseek-r1:7b} (on)), suggesting that most models saturate a similar fraction of the GPU's compute pipeline at peak. The low peak of the smallest models indicates that smaller models never fully utilize the RTX~4060~Ti's shader execution resources.
 
\subsection{Energy Per Prompt}
 
Figure~\ref{fig:gpu_power_per_prompt} plots total GPU energy consumed per prompt. \texttt{qwen2.5:0.5b} achieved the lowest per-prompt energy at \SI{98.6}{\joule}, followed by \texttt{tinyllama:1.1b} at \SI{110.2}{\joule}. The two values are driven by both the lower load power and the higher throughput of these small models, resulting in substantially shorter inference windows.
 
A notable outlier is \texttt{qwen3.5:4b (on)} at \SI{3590.0}{\joule} per prompt --- more than $36\times$ higher than \texttt{qwen2.5:0.5b} --- despite being nominally a 4B-parameter model with load power of only \SI{124}{\watt}. The mean inference duration for this model was \SI{29}{\second} per prompt (compared to \SI{1.6}{\second} for \texttt{qwen2.5:0.5b}). We attribute this to Qwen~3.5's default extended-thinking / chain-of-thought reasoning mode, which generates a large number of internal scratchpad tokens before producing the final response \cite{qwen3}. At \SI{68}{tok\per\second} and a mean inference window of $\approx$\,\SI{29}{\second}, the model generates on the order of \num{1970} tokens per prompt, compared to $\approx$\,\num{515} for \texttt{qwen2.5:0.5b}.
\subsection{Energy Per Output Token}
 
\begin{figure*}[!tb]
 \centering
 \includegraphics[width=\textwidth,trim={0cm 0cm 0cm 0cm},clip]{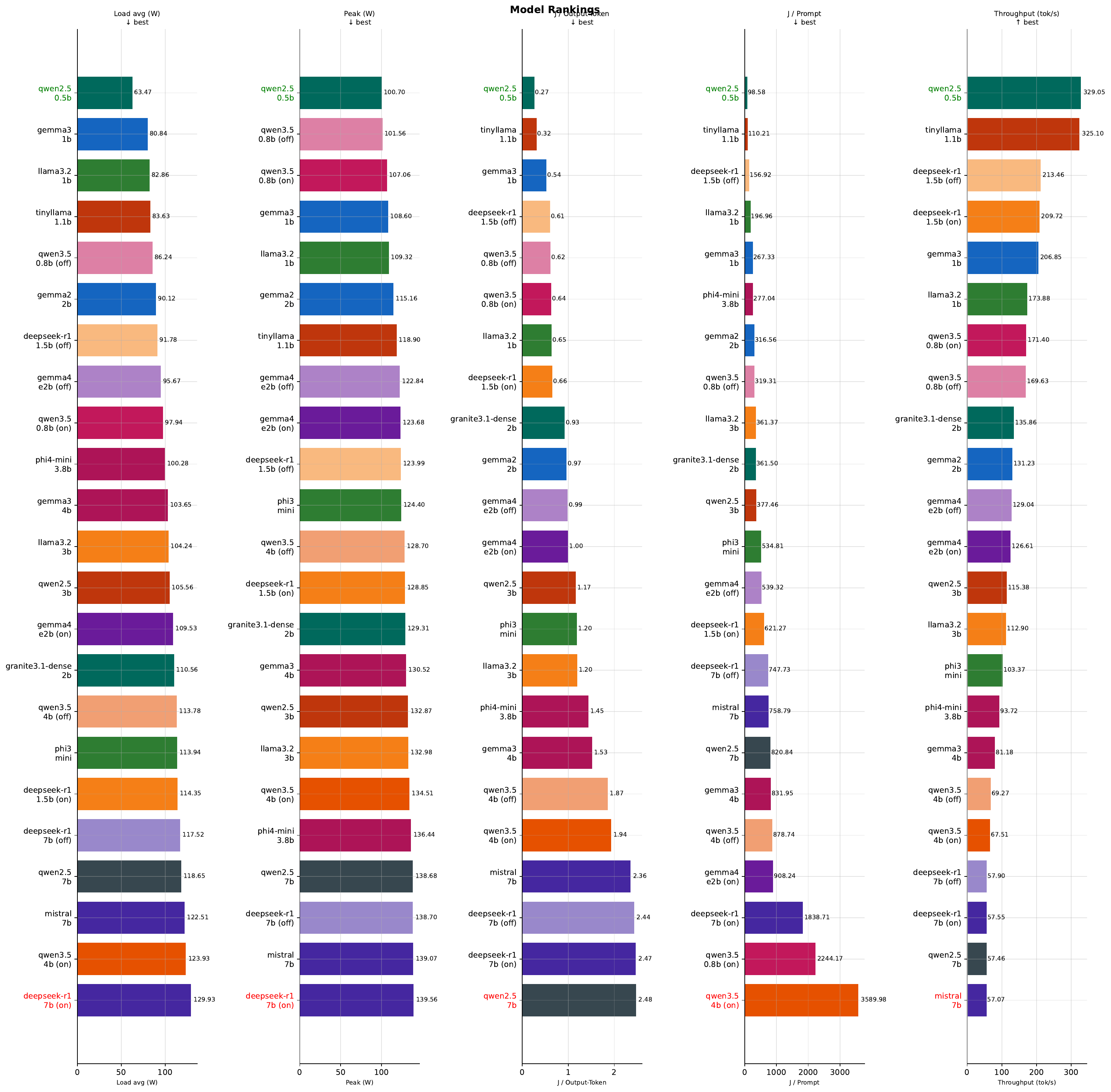}
 \caption{Per-metric model rankings (best to worst). Green labels indicate
      top-ranked models; red labels indicate bottom-ranked models.}
 \label{fig:rankings}
\end{figure*}

Figure~\ref{fig:gpu_power_per_token} and the rightmost ranking column in Figure~\ref{fig:rankings} report J per output token, the primary energy efficiency metric for inference workloads of varying response length.
 
\texttt{qwen2.5:0.5b} ranks first at \SI{0.27}{\joule\per token}, closely followed by \texttt{tinyllama:1.1b} at \SI{0.32}{\joule\per token}. These two models are separated from the rest of the field; the next best model, \texttt{gemma3:1b}, consumes \SI{0.54}{\joule\per token}---approximately 2$\times$ the energy per token of \texttt{qwen2.5:0.5b}.
 
\texttt{qwen2.5:7b} is the least efficient at \SI{2.48}{\joule\per token}, representing a $9.19\times$ penalty relative to the best model. The efficiency gap is not strictly proportional to parameter count: \texttt{phi4-mini:3.8b} (\SI{1.45}{\joule\per token}) consumes notably more energy per token than the 3B-class models \texttt{llama3.2:3b} (\SI{1.2}{\joule\per token}) and \texttt{qwen2.5:3b} (\SI{1.17}{\joule\per token}), suggesting that architectural differences and quantization schemes interact with energy efficiency in non-trivial ways.
 
Interestingly, when evaluated on a per-token basis, \texttt{qwen3.5:4b (off)} (1.87~J/token) and \texttt{qwen3.5:4b (on)} (1.94~J/token) rank substantially closer to the middle of the field than their per-prompt energy would suggest, confirming that their anomalously high per-prompt energy is driven primarily by generating far more tokens per prompt rather than by exceptionally inefficient token generation per se.

\subsection{Inference Throughput}
 
Figure~\ref{fig:rankings} (rightmost panel) ranks models by throughput. \texttt{qwen2.5:0.5b} leads at \SI{329.05}{tok\per\second}, followed closely by \texttt{tinyllama:1.1b} at \SI{325.10}{tok\per\second}. \texttt{gemma3:1b} reaches \SI{206.85}{tok\per\second}.
Throughput degrades monotonically with model size within a family: Gemma~3~1B produces $2.55\times$ more tokens per second than Gemma~3~4B (\SI{81.18}{tok\per\second}). \texttt{mistral:7b} has the lowest throughput at \SI{57.07}{tok\per\second}.
 
The strong inverse relationship between throughput and J/prompt (Pearson $r \approx -0.85$, excluding \texttt{qwen3.5:4b (on)}) highlights that, for tasks where response length is roughly constant, choosing a high-throughput model confers both lower latency and lower total energy costs.
 
\subsection{Thermal Behaviour}
 
\begin{figure*}[!tb]
 \centering
 \includegraphics[width=\textwidth,trim={0cm 0cm 0cm 1cm},clip]{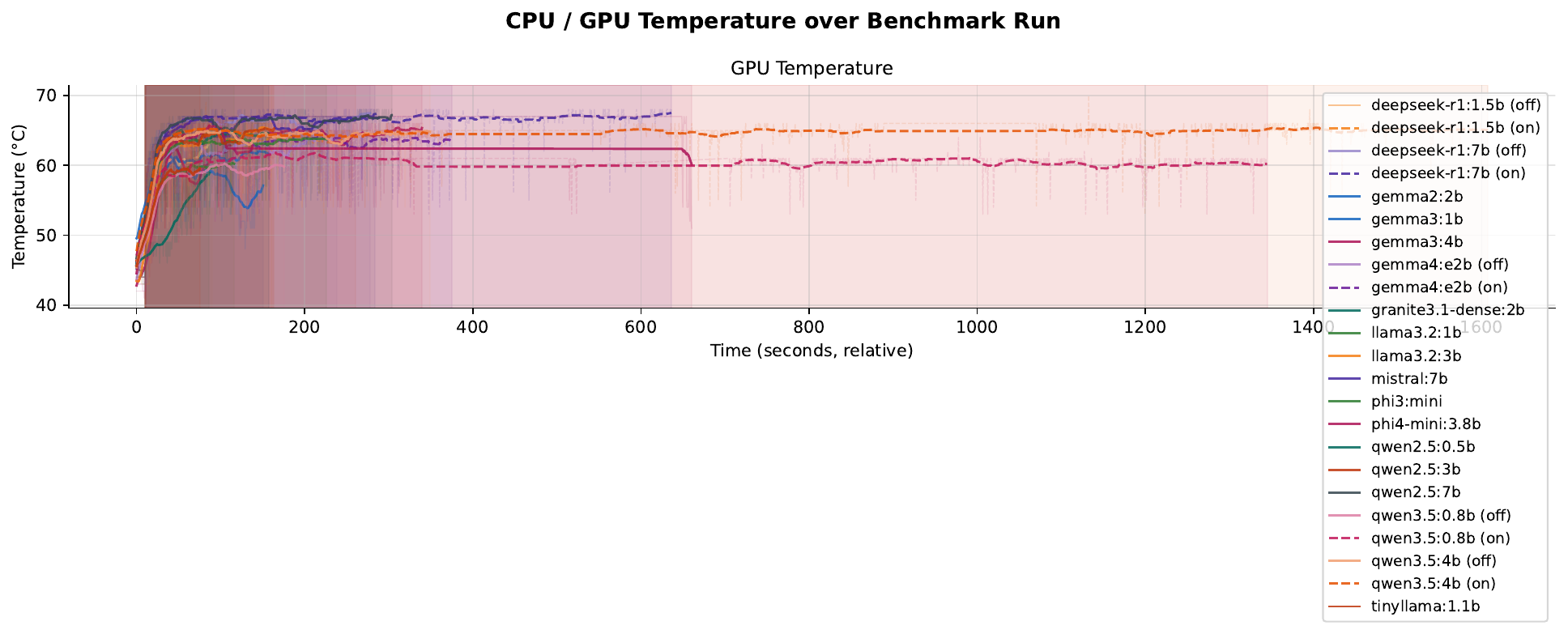}
 \caption{GPU die temperature over the full benchmark run. Shaded regions indicate active inference windows per model.}
 \label{fig:thermal}
\end{figure*}

Figure~\ref{fig:thermal} shows the GPU temperature throughout the full benchmark run. All models caused a rapid temperature rise from idle ($\approx$\,\SI{41}{\celsius}) to a plateau within the first \SI{60}{\second} of inference. Steady-state temperatures ranged from approximately \SI{57}{\celsius} to \SI{65}{\celsius}, all well within the RTX~4060~Ti's thermal envelope \cite{noauthor_nvidia_nodate}. The shaded regions correspond to individual model benchmark windows; the extended pink region visible in the final third of the timeline corresponds to \texttt{qwen3.5:4b (on)}'s prolonged inference period ($>$\,\SI{20}{\minute} total). No thermal throttling events were observed during the benchmark.

\section{Discussion}
\label{sec:discussion}
 
\subsection{Model Size vs.\ Energy Efficiency}
 
Our results confirm that smaller models are more energy-efficient per token, but the relationship is not purely linear. The 3.8B-class \texttt{phi4-mini:3.8b} consumes significantly more energy than both \texttt{llama3.2:3b} and \texttt{qwen2.5:3b}, despite having only marginally more nominal parameters. Architectural features in Phi~4, including mechanisms intended to improve model capability, may therefore come at the cost of higher energy consumption. This observation is consistent with the findings of Argerich and Patiño-Martínez \cite{argerich2024measuring}. Their analysis of the Pythia model family showed that Pythia-1B achieves lower latency and energy consumption per token than the much smaller Pythia-410M model. The authors attribute this to architectural differences: Pythia-410M contains 24 layers, whereas Pythia-1B has only 16. Since layers must be processed sequentially, deeper architectures limit GPU parallelization \cite{argerich2024measuring}. While the underlying causes may differ, these findings illustrate that parameter count alone is not a reliable predictor of inference efficiency. Conversely, the Qwen~2.5 3B model achieves a competitive J/token figure, consistent with the architectural optimizations reported in its technical report~\cite{qwen3}.
 
\subsection{The Hidden Cost of Reasoning Modes}

The \texttt{qwen3.5:0.8b} case study provides a cautionary finding for practitioners: reported model size is an unreliable predictor of per-prompt energy when reasoning-augmented models are involved. The model's default think-before-answer behavior multiplies per-prompt energy by a factor of $\approx$\,10 compared to architecturally similar models. Users deploying such models should either disable extended reasoning where it is unnecessary or explicitly account for the increased token budget in energy projections. This mirrors the analytical prediction of Vartziotis et al. \cite{vartziotis2026tokens}, who show that attention-related KV-cache reads grow quadratically with generated sequence length, becoming the dominant energy term at high output-token counts - consistent with the disproportionate energy cost of reasoning-enabled models observed here.
 
\subsection{Practical Implications}
 
For users running a conversational assistant with approximately 100 interactions per day and an average response of \num{300} output tokens, the annual energy cost difference between the most and least efficient models evaluated here amounts to approximately:
\begin{multline*}
 \Delta E_\mathrm{year} = 100 \cdot 300 \cdot (2.485 - 0.275) \cdot 365 \\
      \approx \SI{24.1}{\mega\joule} \approx \SI{6.7}{\kilo\watt\hour}
\end{multline*}
At a European average residential electricity price of $\approx$\,\EUR{0.30}/kWh~\cite{eurostat2024}, this corresponds to a saving of approximately \EUR{2.01} per year - modest in absolute monetary terms but significant at an organizational scale (e.g., $\approx$\,\SI{67}{\kilo\watt\hour} per 10 users annually). Moreover, the higher-throughput model \texttt{qwen2.5:0.5b} delivers responses in $\approx$\,\SI{0.9}{\second} at 300 tokens vs.\ $\approx$\,\SI{5.4}{\second} for \texttt{mistral:7b}, providing a simultaneous user-experience improvement.
 
\subsection{Limitations}
 
Several limitations of the present study should be acknowledged. First, the benchmark was conducted on a single hardware configuration; results may differ on systems with different CPU-GPU memory bandwidth ratios or on dedicated inference hardware such as Apple Silicon. Second, the prompt set consisted of 15 items stratified across three complexity tiers (short factual, medium reasoning, long generation); while deliberately diverse, it does not constitute a statistically validated evaluation suite for output quality (e.g., MT-Bench, MMLU), meaning that efficiency rankings cannot be directly interpreted as quality-adjusted rankings. Third, \texttt{nvidia-smi}'s \SI{100}{\milli\second} internal averaging window and the \SI{500}{\milli\second} polling interval may underestimate instantaneous peak power for very short inference calls. Fourth, the Ollama runtime includes Python/HTTP overhead that is not separately accounted for; CPU energy contributions are not measured. Finally, the default quantization configurations shipped with Ollama are used, so custom quantization schemes could yield different results.
 
\subsection{Future Work}
 
Future work should (i) extend the benchmark to ARM-based SoCs (Apple M-series, Qualcomm Snapdragon X), where power measurement APIs differ substantially; (ii) incorporate a quality metric such as MT-Bench or MMLU to enable energy-quality Pareto analysis; (iii) instrument CPU, DRAM, and peripheral power to obtain full system energy rather than GPU-only figures; (iv) investigate the interaction between quantization bit-width, batch size, and per-token energy to guide deployment decisions; and (v) calibrate analytical estimators such as that of Vartziotis et al. \cite{vartziotis2026tokens} against consumer-GPU measurements to test whether H100-derived hardware coefficients ($\alpha_{\text{TC}}$, $e_{\text{HBM}}$) generalize to Ada Lovelace-class accelerators.

\section{Conclusion}
\label{sec:conclusion}
 
We presented a systematic, hardware-level energy benchmark comparing eighteen openly available language models on a consumer-grade GPU. Our results demonstrate that \texttt{qwen2.5:0.5b} and \texttt{tinyllama:1.1b} dominate the energy-efficiency frontier, delivering the lowest energy per output token (\SI{0.27}{} and \SI{0.32}{\joule\per token}, respectively) and the highest throughput ($>$\,\SI{325}{tok\per\second}). The 7B Mistral model consumes $8.6\times$ more energy per token, confirming that model scale carries a substantial energy penalty even on dedicated consumer hardware.
 
A key qualitative finding is that energy efficiency is not a simple function of parameter count: architectural family, quantization strategy, and especially the presence of reasoning-augmented generation modes can dominate energy profiles and must be considered independently.
 
We release all benchmark code and data openly to support reproducible research in this area. As locally deployed LLMs become ubiquitous, incorporating energy metrics alongside accuracy and latency in model selection decisions will be essential for responsible, sustainable AI deployment.